\documentclass[runningheads]{llncs}
\usepackage[T1]{fontenc}
\usepackage{graphicx}
\usepackage{amsmath}
\usepackage{float}
\usepackage[utf8]{inputenc}
\usepackage{amssymb}
\usepackage[square,numbers]{natbib}
\usepackage{siunitx}    
\usepackage{booktabs}   
\usepackage{multirow}

\usepackage[table]{xcolor}
\usepackage{amsmath}
\usepackage[ruled,vlined,linesnumbered]{algorithm2e}
\usepackage{xcolor}
\usepackage{enumitem}
\usepackage{hyperref}
\usepackage{tabularx}
\usepackage{orcidlink}
\usepackage{balance}
\usepackage{colortbl}
\usepackage{tabularx, booktabs}
\renewcommand{\orcidID}[1]{\orcidlink{#1}}
\DontPrintSemicolon
\SetKwInput{Input}{Input}
\SetKwInput{Output}{Output}
\SetKwComment{tcc}{\color{gray}\# }{}
\SetKw{Return}{return}
\SetKwBlock{Repeat}{repeat}{until convergence}

\newcommand{\theName}{{DeLIVeR}}

\usepackage{eso-pic}

\begin{document}

\AddToShipoutPictureBG*{%
  \AtPageUpperLeft{%
    \put(\LenToUnit{0.5\paperwidth},\LenToUnit{-2.0cm}){%
      \makebox[0pt][c]{%
        \begin{minipage}{0.7\paperwidth}
          \centering \footnotesize
          Cong Hoan Nguyen, Thomas Hoang, Minh Hieu Duong, and Long Nguyen. ``DeLIVeR: Decomposed Learning for Information-grounded Veracity Recognition via Reinforced Knowledge Graph Exploration'', 7th International Conference on Deep Learning Theory and Applications (DeLTA 2026).
        \end{minipage}%
      }
    }
  }
  \AtPageLowerLeft{%
    \put(\LenToUnit{0.5\paperwidth},\LenToUnit{2.0cm}){%
      \makebox[0pt][c]{%
        \begin{minipage}{0.7\paperwidth}
          \centering \scriptsize
          \copyright 2026 Springer. Personal use of this material is permitted. Permission from Springer must be obtained for all other uses, in any current or future media, including reprinting/republishing this material for advertising or promotional purposes, creating new collective works, for resale or redistribution to servers or lists, or reuse of any copyrighted component of this work in other works.
        \end{minipage}%
      }
    }
  }
}

\title{DeLIVeR: Decomposed Learning for Information-grounded Veracity Recognition via Reinforced Knowledge Graph Exploration}

\titlerunning{Decomposed Learning for Information-grounded Veracity Recognition}

\author{Cong Hoan Nguyen\inst{1}\orcidID{0009-0007-5046-1453} \and
Thomas Hoang\inst{2}\orcidID{0009-0001-5749-8837} \and
Minh Hieu Duong\inst{1}\orcidID{0000-0002-4715-8169}\and
Long Nguyen\inst{1}\orcidID{0000-0001-7673-7955}}


\institute{University of Louisville, Louisville KY 40292, USA.\\
\email{\{conghoan.nguyen,hieu.duong,l.nguyen\}@louisville.edu}\\ \and
Denison University, Granville, Ohio 43023, USA.\\
\email{\{hoang\_t2\}@denison.edu}}

\maketitle              

\begin{abstract}

Automated fact-checking remains a challenge for Large Language Models (LLMs) due to "query brittleness" in traditional retrieval systems. We propose DeLIVeR (\textbf{De}composed \textbf{L}earning for \textbf{I}nformation-grounded \textbf{Ve}acity \textbf{R}ecognition), a framework that treats evidence retrieval as a reinforced strategic exploration task. \theName\ utilizes a Planner LLM to decompose complex claims into targeted question sets, which are used to traverse structured Knowledge Graphs (KGs) for high-precision evidence. We optimize the Planner’s policy using Group Relative Policy Optimization (GRPO) with a reward system prioritizing structural diversity and verdict accuracy.
Our evaluation on LIAR, FEVER, and PolitiFact shows that \theName\ significantly outperforms state-of-the-art baselines. Using Qwen2.5-7B, our framework achieved peak F1-scores of 83.73, 84.57, and 79.70 respectively, representing a 10–15\% improvement over HippoRAG2. By shifting to a reinforced question-planning strategy, \theName\  effectively bridges multi-hop reasoning gaps and provides an auditable, transparent path for verifiable misinformation detection.

\end{abstract}

\section{Introduction}

The proliferation of online misinformation poses a critical threat to public trust and democratic stability. While Large Language Models (LLMs) offer sophisticated reasoning, they are frequently undermined by ``hallucinations'' stemming from a reliance on static internal knowledge rather than verifiable grounding \cite{ji2023survey, zhang2025siren}. Although Retrieval-Augmented Generation (RAG) aims to mitigate this, existing frameworks struggle with \textbf{multi-hop gaps}, \textbf{static query limitations}, and \textbf{fixed retrieval policies} that fail to adapt to complex claims spanning multiple entities \cite{lewis2020retrieval, potthast2018stylometric, qian2018neural}.

\begin{figure}[htbp]
\centering
\includegraphics[width=\linewidth]{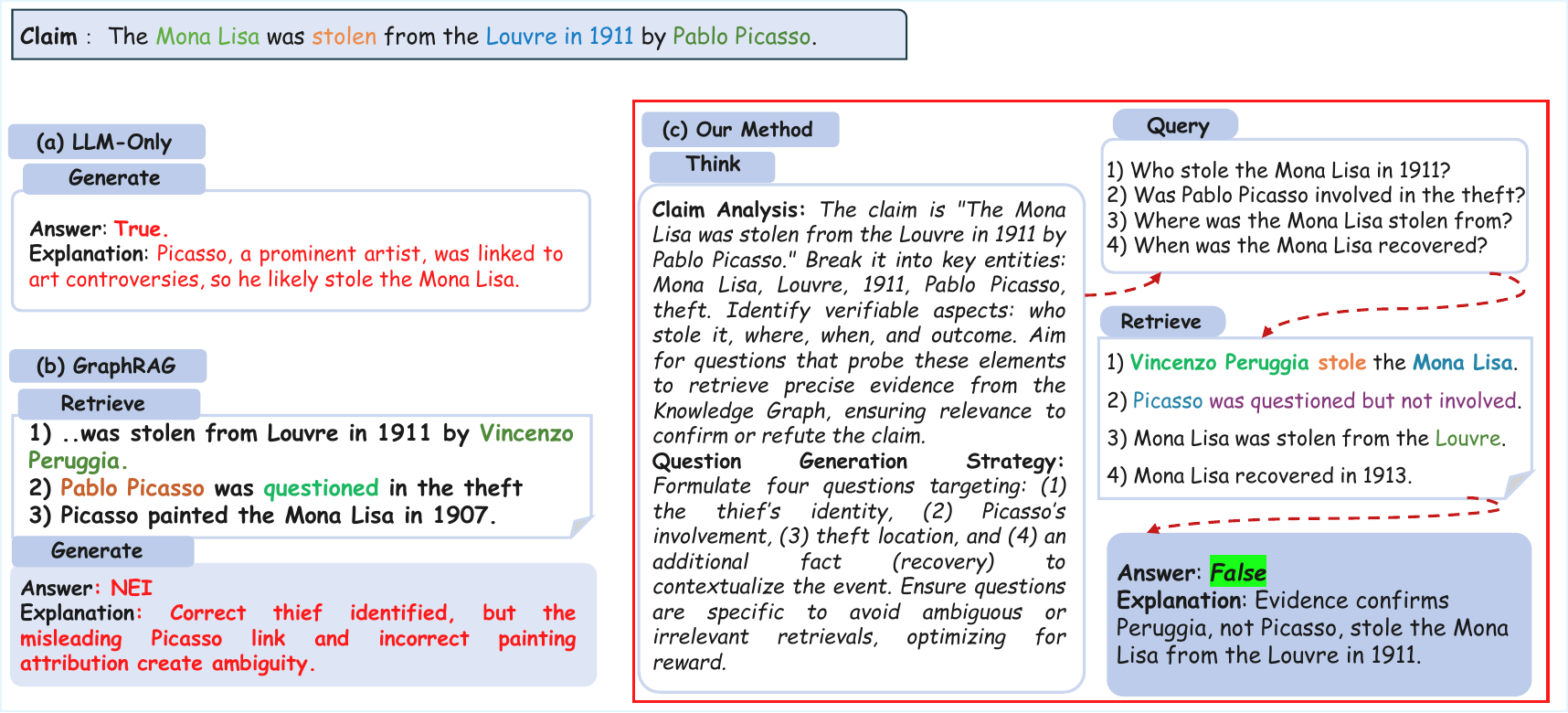}
\caption{Comparison of fake news detection systems for the Mona Lisa theft claim.}
\label{fig:introduction}
\end{figure}

To address these limitations, we propose \theName\, a framework that transforms fact verification from a static retrieval step into a reinforced, closed-loop optimization process as shown in Figure \ref{fig:introduction}. Unlike domain-adaptive approaches that focus primarily on feature-level transformations, \textsc{DeLIVeR} optimizes the model's information-seeking strategy. Given a claim, a planner LLM generates a small set of diverse, targeted questions that collectively query a Knowledge Graph (KG) to retrieve complementary evidence. This policy is optimized via Group Relative Policy Optimization (GRPO) to reward structural diversity and verdict accuracy. In developing this framework, we specifically investigate whether a reinforced planner LLM can effectively bridge multi-hop reasoning gaps compared to static RAG baselines \textbf{(RQ1)}, the extent to which set-based question planning reduces query ambiguity in structured KGs \textbf{(RQ2)}, and if GRPO-driven optimization yields a more stable and interpretable information-seeking policy for veracity classification \textbf{(RQ3)}.

\textbf{Our contributions are as follows:}
\begin{itemize}
  \item \textbf{Structured multi-hop grounding:} We introduce a KG-grounded evidence retrieval pipeline that moves beyond flat text retrieval by retrieving structured, multi-hop evidence paths for claim verification.
  \item \textbf{Set-based question planning:} We design a planner LLM that decomposes each claim into a cohesive set of diverse, targeted questions, reducing query ambiguity and improving evidence coverage over standard single-query RAG.
  \item \textbf{GRPO-driven policy optimization:} We fine-tune the question generation module with GRPO using rewards that encourage valid format, structural diversity, and evidence quality, yielding a more stable information-seeking policy.
  \item \textbf{Empirical validation and interpretability:} Experiments on FEVER, LIAR, and PolitiFact show consistent improvements over strong RAG baselines, while the question-driven retrieval process provides inherent evidence chains for auditing model decisions.
\end{itemize}

\section{Related Work}

Early artificial intelligence systems for knowledge-intensive tasks relied heavily on neural architectures such as Artificial Neural Networks (ANNs) and energy-based models, particularly Discrete Hopfield Neural Networks (DHNNs), which demonstrated the foundational principles of structured reasoning and pattern recognition. These early approaches established critical insights into logic mining and constraint satisfaction that continue to inform modern AI systems. Notable contributions include flexible logic mining frameworks that combine ensemble multi-attribute selection with DHNNs \cite{gao2026fgra}, investigations into weighted C-type random 2-satisfiability formulations within discrete Hopfield networks \cite{chang2025weighted}, and optimized logic mining methods leveraging higher-order satisfiability representations \cite{romli2025optimized}. While these early neural and logic-based paradigms demonstrated that structured, rule-governed representations significantly improve classification reliability and interpretability, they typically operate on fixed propositional representations and do not address the open-domain, multi-hop evidence retrieval needed for modern fact verification over large, dynamic knowledge corpora.

Fake news detection has progressed from feature-based models \citep{shu2017fake, turchi2015mt} to pretrained deep encoders such as BERT and RoBERTa \cite{devlin2019bert, liu2019roberta}, which perform well on benchmarks including LIAR \cite{wang2017liar} and FEVER \cite{thorne2018fever}. However, veracity prediction with LLMs remains unreliable without explicit grounding, due to hallucinations and the absence of verifiable evidence attribution \cite{zhang2023towards}. Retrieval-Augmented Generation (RAG) addresses this by conditioning predictions on external context \cite{lewis2020retrieval}, but standard RAG typically retrieves unstructured text using similarity-based queries and often fails to capture the relational structure needed for claims involving multiple entities and linked events \cite{jiang2023active, khattab2022demonstrate}. Recent graph-based retrieval methods improve indexing and ranking, but they commonly treat the user query as fixed. In contrast, \theName\ emphasizes structured KG evidence while optimizing the information-seeking strategy through learned question sets.

A second challenge is query ambiguity. Many RAG systems rely on a single query derived directly from the claim, making retrieval brittle when the claim is underspecified or implicitly framed. Prior work on claim decomposition shows that generating intermediate questions can improve reasoning \cite{press2023measuring, chen2022generating}, and question generation models such as T5/BART have been used to produce subquestions \cite{ousidhoum2022varifocal}. However, these approaches are often not trained to optimize the retrieval outcome needed for verification, especially when evidence must be gathered from complementary semantic facets (e.g., temporal context, relations, contradictions). Our approach introduces a Planner LLM that generates a small, diverse set of targeted questions to probe a KG from multiple angles, improving evidence coverage and reducing ambiguity for downstream verification \cite{yasunaga2021qa}.

Finally, optimizing retrieval behavior for black-box LLM pipelines is difficult because gradients cannot be propagated through the verifier. Existing methods align retrieval with LLM feedback using objectives such as KLD-based techniques \cite{shi2024replug} or policy-gradient style optimization \cite{ouyang2022training}, but PPO-style approaches can be unstable in high-dimensional generation spaces \cite{schulman2017proximal}. We address this by applying Group Relative Policy Optimization (GRPO) \cite{shao2024deepseekmath} to optimize question-set generation. GRPO estimates advantages by comparing multiple outputs from the same claim, enabling stable updates without a centralized critic. This provides a practical mechanism to learn a robust question-generation policy that improves retrieval precision and supports accurate veracity decisions by rewarding high-signal evidence while discouraging noisy retrieval.

\section{Methodology}

\subsection{Overview}

\begin{figure}[h]
\centering
  \includegraphics[width=0.9\textwidth]{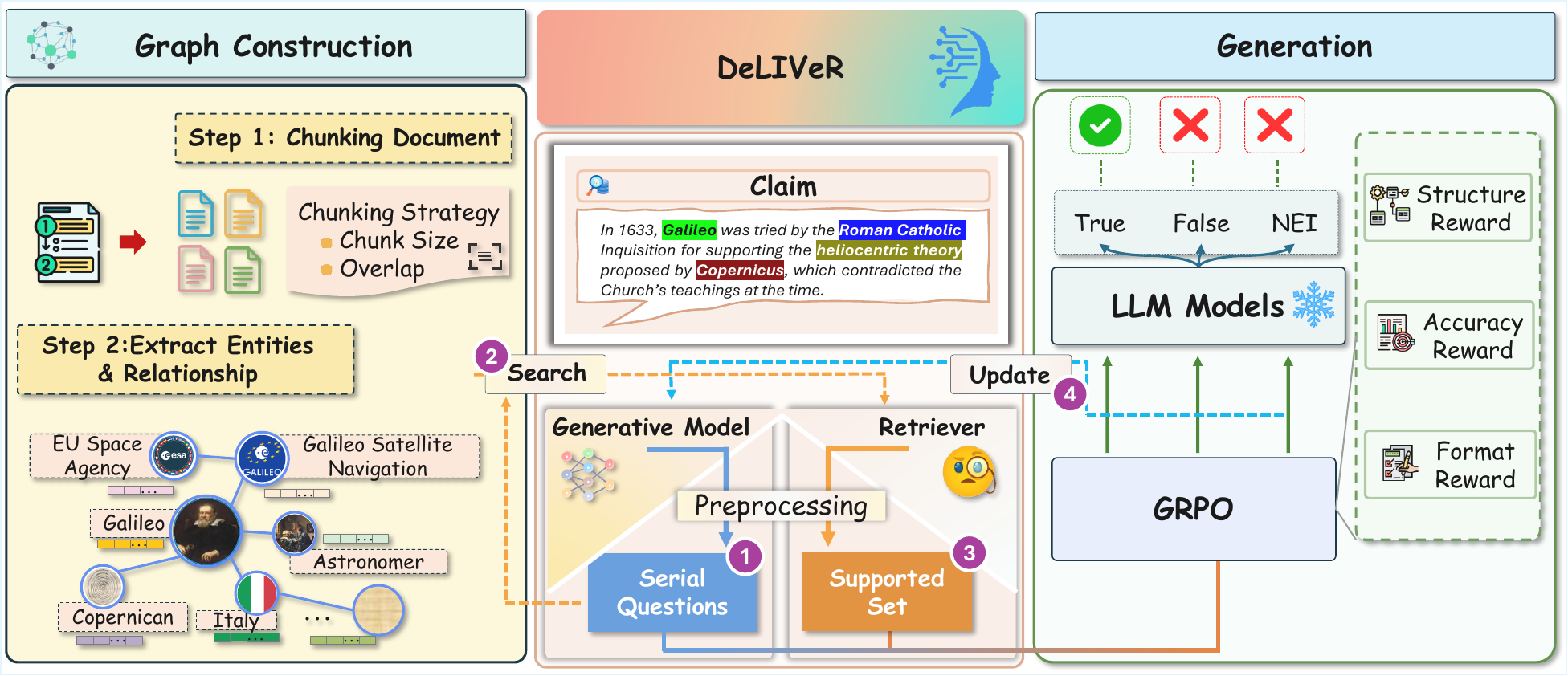}
  \caption{The overall \theName\ architecture operates through four key stages: generating a sequence of questions, querying the knowledge graph, extracting the supporting information set, and updating the generative model using GRPO.}
  \label{fig:teaser}
\end{figure}

The proposed \theName\  framework presented in Figure \ref{fig:teaser} consists of Knowledge Graphs (KGs), a two footprint LLM architecture and reinforcement learning to tackle the challenges associated with fake news detection by accurate evidence retrieval and knowledge-based generating a verdict. The system begins with KG sourced from a reliable dataset, with nodes representing entities or facts, with edges indicating relationships (for example "supports", "contradicts", etc), allowing for structured and contextual evidence query \cite{hogan2021knowledge}. 
The secondary LLM, fine-tuned for question generation, processes the input claim to produce a cohesive set of targeted questions, collectively querying the KG to retrieve comprehensive evidence \citep{lewis2019bart}. The aggregated evidence from the question set, along with the claim, is then fed into a frozen primary LLM to generate a verdict (True, False, Not Enough Information) with a human-interpretable explanation, ensuring stability and interpretability \citep{lewis2020retrieval}. Group Relative Policy Optimization (GRPO) iteratively refines the question-generating LLM by evaluating the quality of retrieved evidence (e.g., format, structure, accuracy), optimizing retrieval effectiveness through reinforced feedback \citep{ouyang2022training}.

This framework integrates other ideas previously proposed to produce an approach that integrates structured knowledge, adaptive queries, and reinforced feedback to provide robust and explainable means of detecting fake news.

\subsection{Problem Formulation}
The veracity detection task in \textsc{DeLIVeR} is governed by the following conditional probability distribution, which decomposes the verdict $y$ into a sequence of planning and retrieval steps:
\begin{equation}
    P(y | c) \approx \max_{\mathcal{Q}} \left[ P(y | c, \text{Retrieve}(\mathcal{Q}, \mathcal{G})) \cdot P(\mathcal{Q} | c; \theta_Q) \right]
\end{equation}
where $c$ is the input claim, $\mathcal{Q} = \{q_1, \dots, q_n\}$ is the set of generated questions, and $\mathcal{G}$ represents the structured Knowledge Graph. Our objective is to optimize the parameters $\theta_Q$ such that the generated $\mathcal{Q}$ maximizes the likelihood of the correct verdict $y$ through high-quality evidence retrieval.

\subsection{Knowledge Graph Construction}
To ensure the framework is grounded in verifiable facts while strictly avoiding label leakage and circular reasoning, we construct the Knowledge Graph (KG) using only the ground-truth evidence corpora associated with each dataset. We explicitly exclude claim text and veracity labels from the graph construction pipeline, utilizing GPT-4 to perform Open Information Extraction (OpenIE) that transforms unstructured evidence into structured triples $(s, p, o)$ (Subject–Predicate–Object). To guarantee high-fidelity retrieval, we implement a provenance-tracking mechanism that maps every node and edge back to its source document URI, allowing the Verifier LLM to audit retrieved paths against the original text. 

\begin{table}[ht]
\centering
\caption{Quality Assessment of Extracting KG Triples}
\label{tab:kg-quality}
\begin{tabular}{l r l}
\toprule
\textbf{Metric} & \textbf{Score} & \textbf{Definition} \\
\midrule
Entity Precision   & 94.2\% & Accuracy of extracted subjects and objects against source text. \\
Relation Recall    & 88.5\% & Percentage of key relations from text captured in the graph. \\
Triple Fidelity    & 91.8\% & Correctness of the (s, p, o) link as a logical unit. \\
Noise Rate         & 3.4\%  & Percentage of extracted triples with no basis in source text. \\
\bottomrule
\end{tabular}
\end{table}

As demonstrated in Table \ref{tab:kg-quality}, a manual quality assessment of 500 randomly sampled triples confirms the reliability of this process, yielding an Entity Precision of 94.2\% and a Triple Fidelity of 91.8\%. Furthermore, we address temporal sensitivity by appending temporal metadata (timestamps) to edges when available in the source text. During retrieval, the model prioritizes edges with timestamps closest to the claim's publication date. For noise control, we apply a frequency-based filter that removes "singleton" entities that do not connect to at least two other nodes, ensuring the graph focuses on the dense, multi-hop relationship clusters necessary for complex reasoning.

\subsection{Question Generation Module}

The Question Generation Module is a core part of the \theName\  framework that leverages a hundred million parameters pretrained secondary large language model, to create relevant questions based on an input claim, in order to frame the language for targeted retrieval of relevant evidence from the KG for fake news verification. Formally, given an input claim \( c \), the secondary LLM $Q$ generates a set of questions \( Q = \{q_1, q_2, \ldots, q_n\} \), with each question $q_i$ to explore specific angles of the claim such as the source, evidence to substantiate claim $c$, contextual factors which may affect the claim, or counter claims or contradictions, such as! “What is the original source of claim $c$?” “Is evidence $x$ in the KG contradictory to assertion $y$ in claim $c$?” “What time factors impact the validity of event $z$?”. The process is defined as \( \mathcal{Q}(c; \theta_Q) \to Q \subseteq \mathcal{P}(\text{KG}) \), where \( \theta_Q \) are the pretrained parameters of the LLM, and the question set \( Q \) is used collectively to query the KG, retrieving a unified set of evidence. Each question generated is intended to cover multiple angles of the claim, such as factual, temporal, and causal, to facilitate a more richly supported collection of evidence from the KG and support multi-hop reasoning, which is necessary in the evaluation of complex misinformation. The question set is mapped to KG nodes and edges using vector similarity search, with embeddings of questions and KG elements ensuring contextually relevant evidence retrieval:
\begin{equation}
\text{Score}(q_i, e_j) = \cos(\phi(q_i), \phi(e_j)),    
\end{equation}
where \( \phi(q_i) \) and \( \phi(e_j) \) are embeddings for the question and KG element (node or edge) \( e_j \), respectively, and \( \cos \) denotes cosine similarity \citep{reimers2019sentence}. This scoring enables the selection of the most relevant KG elements, significantly improving the granularity and relevance of retrieved evidence compared to static query methods in standard RAG systems.

To address redundancy and ensure comprehensive coverage of the KG’s relational structure, the question generation process incorporates a diversity constraint, encouraging the LLM to produce questions that span distinct subgraphs of the KG. This is formalized by maximizing the entropy of question coverage:

\begin{equation}
H(Q) = -\sum_{q_i \in Q} P(q_i | c) \log P(q_i | c),    
\end{equation}
where \( P(q_i | c) \) is the probability of generating question \( q_i \) given claim \( c \), ensuring that questions probe varied aspects of the KG without overlap \citep{yang2018hotpotqa}.

The generated questions are mapped to KG nodes and edges using a vector similarity search. We utilize the embeddings of both the questions and the KG elements to generate evidence that is contextually relevant to the nuance of the claims. The performance of the set of questions is refined iteratively using Group Relative Policy Optimization (GRPO), detailed in Section \ref{sec:grpo}, which adjusts \( \theta_Q \) by evaluating the quality of the retrieved aggregated evidence. This approach, different from static RAG pipelines, optimizes question sets to maximize the extraction of relevant evidence, enhancing the verification of abstract or ambiguous claims \citep{ouyang2022training}. In practice, this modification of objective function is what separates our approach from static RAG pipelines, where retrieving incomplete or irrelevant evidence is difficult. By incorporating pre-trained LLMs with KG retrieval optimized by GRPO, the Question Generation Module provides a necessary level of contextualized evidence extraction to support the system’s verification capabilities for complex claims in fake news detection and misinformation combat.

\label{sec:grpo}

\subsection{Retrieval and Augmentation}

The Retrieval and Augmentation module leverages a cohesive set of questions generated by the secondary LLM to collectively query the Knowledge Graph (KG) and retrieve a unified set of relevant evidence, which is then augmented with the input claim to enable the primary LLM to generate accurate verdicts for fake news detection. For a given claim \( c \) and its set of questions \( Q = \{q_1, q_2, \ldots, q_n\} \), the questions are embedded using a pretrained sentence encoder, and the evidence is retrieved from the KG by computing similarity scores between the question embeddings and the KG node/edge embeddings, formalized as: 
\begin{equation}
\text{Retrieve}(Q, G) = \bigcup_{q_i \in Q} \arg\max_{e_j \in G} \cos(\phi(q_i), \phi(e_j)),
\end{equation}
where \( G = (V, E) \) is the KG, \( \phi(\cdot) \) denotes the Sentence-BERT embeddings, and \( \cos \) is cosine similarity \citep{reimers2019sentence}. The aggregated evidence set \( E = \{e_1, e_2, \ldots, e_m\} \) is concatenated with the claim \( c \) to form an augmented input \( [c; E] \), which is fed to the primary LLM to produce a verdict (True, False, or Not Enough Information) with an explanation. The quality of the retrieved evidence is evaluated to inform iterative refinement of the question generation process \citep{lewis2020retrieval}. This process ensures the primary LLM leverages structured, contextually relevant evidence, enhancing verdict accuracy and interpretability compared to standard RAG systems that rely on unstructured text retrieval.

\begin{figure}[h]
\centering
\includegraphics[width=0.8\linewidth]{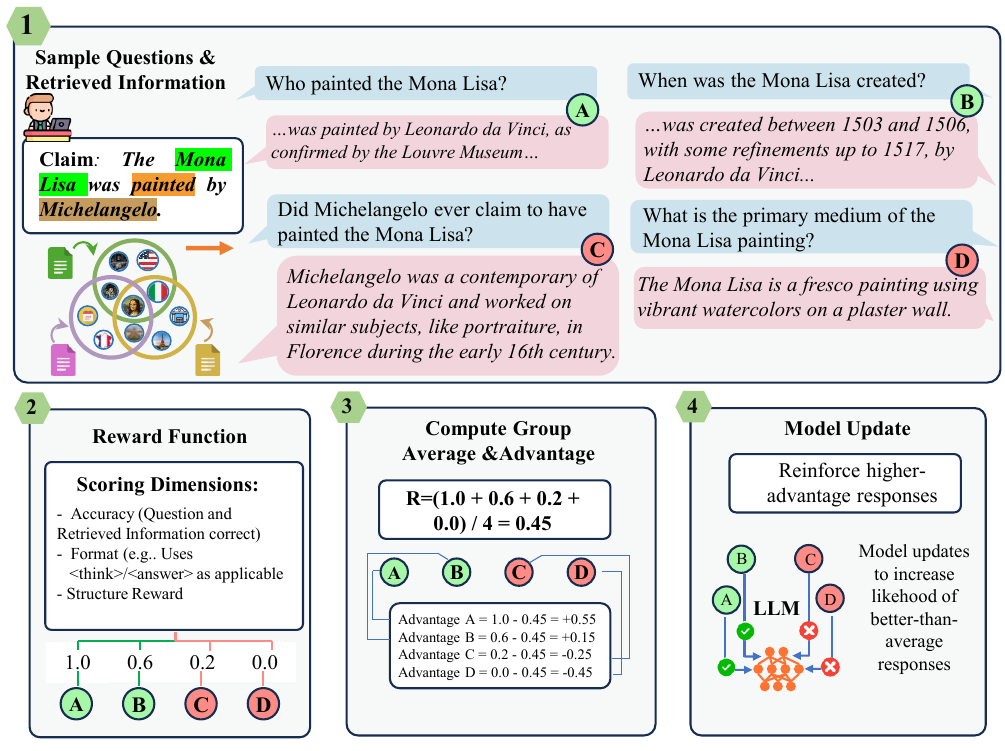}

\caption{Overview of the Group Relative Policy Optimization (GRPO) process for refining question generation. Step 1: samples question sets \( \{Q_i\} \) from the policy \( \pi_{\theta_Q} \) and retrieves corresponding evidence \( E_i \) from the Knowledge Graph. Step 2: computes the Format-Structure-Accuracy reward \( R(Q_i, E_i) \) for each set. Step 3: calculates the group average reward and advantage \( \hat{A}(Q_i, E_i) \). Step 4: updates the policy parameters \( \theta_Q \) via the clipped objective to optimize retrieval effectiveness and verdict reliability.}
\label{fig:Overall_GRPO_process}
\end{figure}

\subsection{Reinforcement Learning Optimization}

To bridge the gap between static retrieval and adaptive fact-checking, we optimize the Question Generation (QG) module using Group Relative Policy Optimization (GRPO) \citep{shao2024deepseekmath}. Unlike standard Proximal Policy Optimization (PPO), which relies on a centralized critic to estimate a state-value baseline, GRPO computes advantages based on the relative performance of a group of outputs generated from the same prompt. This is particularly advantageous for our framework because it allows the model to compare multiple diverse "question sets" for a single claim, identifying which specific combinations of queries maximize evidence coverage while minimizing retrieval noise.
Additionally, this approach enhances the quality of evidence retrieved from the Knowledge Graph (KG) for fake news detection, as illustrated in the GRPO process overview (see Figure \ref{fig:Overall_GRPO_process}). Given a claim \( c \in \mathcal{D} \), the LLM generates a group of question sets \( \{Q_i\}_{i=1}^N \subseteq \mathcal{T}_c \), where each \( Q_i = \{q_{i1}, \ldots, q_{iT}\} \) is used to collectively query the KG, producing an aggregated evidence set \( E_i = \text{Retrieve}(Q_i, G) \). We use the prompt, detailed in Table \ref{tab:grpo_prompt}, optimize the policy \( \pi_{\theta_Q} \) using the GRPO objective, defined as:
{\footnotesize
\begin{equation}
\begin{aligned}
J_{\text{GRPO}}(\theta_Q) = \mathbb{E}_{c \sim \mathcal{D}, \{Q_i\}_{i=1}^N \sim \pi_{\theta_Q^{\text{old}}}(Q|c;\text{KG})} \Bigg[ \frac{1}{N} \sum_{i=1}^N \min \Big( \rho_{\theta_Q}(Q_i) \hat{A}(Q_i, E_i), \\
\text{clip}\left( \rho_{\theta_Q}(Q_i), 1 \pm \epsilon \right) \hat{A}(Q_i, E_i) \Big) - \beta D_{\text{KL}}(\pi_{\theta_Q} \parallel \pi_{\text{ref}}) \Bigg],
\end{aligned}
\end{equation}
}
where \( \rho_{\theta_Q}(Q_i) = \frac{\pi_{\theta_Q}(Q_i | c; \text{KG})}{\pi_{\theta_Q^{\text{old}}}(Q_i | c; \text{KG})} \), and the advantage is:
$\hat{A}(Q_i, E_i) = R(Q_i, E_i) - \frac{1}{N} \sum_{j=1}^N R(Q_j, E_j) / F_{\text{norm}}(\{R(Q_j, E_j)\}_{j=1}^N)$
and the reward function:
$R(Q_i, E_i) = w_1 \cdot R_{\text{Format}}(Q_i) + w_2 \cdot R_{\text{Structure}}(Q_i) + w_3 \cdot R_{\text{Accuracy}}(E_i),$
where \( F_{\text{norm}}(\cdot) \) is a normalization function (e.g., standard deviation), and \( \text{clip}(\cdot) \) stabilizes policy updates \citep{schulman2017proximal}. GRPO’s group-based comparison optimizes question sets to maximize the quality of retrieved evidence, enhancing retrieval effectiveness and verdict reliability over traditional RL methods \citep{lewis2020retrieval, ouyang2022training}.

\begin{table}[!h]
\centering
\scriptsize
\renewcommand{\arraystretch}{1.3}
\begin{tabular}{p{\dimexpr\columnwidth-2\tabcolsep-2\arrayrulewidth}}
\hline
\rowcolor{cyan!20}\textbf{Prompt} \\
\hline
A conversation between User and Assistant for training a model to generate questions from a claim. The User provides a claim, and the Assistant generates 4–8 relevant questions to query a Knowledge Graph or external sources. The Assistant outlines the reasoning process within \textcolor{blue}{ \textbf{ \texttt{<think>...} \texttt{</think>}}} tags, detailing how the claim is parsed and questions are formulated, then provides the questions within \textcolor{orange}{ \textbf{ \texttt{<answer>...} \texttt{</answer>}}} tags. 

\vspace{0.5em}
\textbf{User:} Given a claim, generate 4–8 diverse and specific questions to query a Knowledge Graph or external sources for misinformation detection. Show your reasoning process and list the questions. 

\vspace{0.5em}
\textbf{Assistant:} 

\textcolor{blue}{ \textbf{\texttt{<think>}}} 

1. \textit{Parse the claim:} Identify key entities, events, relationships, and context within the claim to ensure questions target critical aspects. 

2. \textit{Formulate questions:} Generate 4 questions that are diverse, covering different facets (e.g., factual details, entities, temporal aspects, contradictions), and specific to maximize relevance for Knowledge Graph retrieval or external search. Ensure questions are clear, concise, and optimized for misinformation detection on datasets. 

3. \textit{Optimize diversity:} Balance question types (e.g., who, what, when, where, why) to cover multiple perspectives while avoiding redundancy. 

\textcolor{blue}{ \textbf{\texttt{</think>}}} 
\vspace{0.5em}

\textcolor{orange}{ \textbf{\texttt{<answer>} }}

List of 4 questions: 

\textcolor{blue}{ \textbf{\texttt{<question 1>}}} \textcolor{blue}{ \textbf{\texttt{</question 1>}}} \\
\textcolor{blue}{ \textbf{\texttt{<question 2>}}} \textcolor{blue}{ \textbf{\texttt{</question 2>}}} \\
\textcolor{blue}{ \textbf{\texttt{<question 3>}}} \textcolor{blue}{ \textbf{\texttt{</question 3>}}} \\
\textcolor{blue}{ \textbf{\texttt{<question 4>}}} \textcolor{blue}{ \textbf{\texttt{</question 4>}}} \\

\textcolor{orange}{ \textbf{\texttt{</answer>}}} 
\\

Claim: \textcolor{red}{claim}. Assistant:
\\  
\hline
\caption{Prompt template for \theName\  question generation \label{tab:grpo_prompt}}
\end{tabular}
\end{table}

\noindent
\textbf{Format Reward (\( R_{\text{Format}}(Q_i) \))}.  
This reward enforces strict adherence to the structured output format required for downstream retrieval and parsing. The model must generate a reasoning trace within \textbf{\texttt{<think>} \texttt{</think>}} tags followed by a numbered list of 4--8 questions within \texttt{<answer>} \texttt{</answer>}, with no extraneous text. We use regular expressions to validate:  
- Exactly one \texttt{<think>} block with coherent reasoning steps.  
- Exactly one \texttt{<answer>} block containing a numbered list (\texttt{1.}, \texttt{2.}, etc.) with 4 $\leq$ questions $\leq$ 8.  
- No content outside tags.  
This is a \textbf{binary reward}: 

\begin{equation}
R_{\text{Format}}(Q_i) =
\begin{cases}
1 & \text{if format is valid}, \\
0 & \text{otherwise}.
\end{cases}
\end{equation}

\noindent
\textbf{Structural Reward (\( R_{\text{Structure}}(Q_i) \))}.  
This is the \textbf{core signal} for effective evidence retrieval. It evaluates whether the generated question set \( Q_i \) covers \textbf{diverse semantic dimensions} of the claim \( c \) (e.g., entities, events, temporal context, causal links, contradictions) to maximize informational breadth. The computation proceeds in two steps:  
1. Each question \( q_{ij} \in Q_i \) is classified into one of \( K \) predefined \textbf{semantic categories} (e.g., \texttt{WHO}, \texttt{WHAT}, \texttt{WHEN}, \texttt{WHERE}, \texttt{HOW}, \texttt{CONTRADICTION}) using a lightweight classifier (e.g., fine-tuned BERT or keyword patterns). Let \( C(Q_i) \) be the set of unique categories covered.  
2. The reward is the \textbf{coverage ratio} relative to an ideal category distribution \( C^* \) (derived from claim type or oracle analysis):  
\begin{equation}
    R_{\text{Structure}}(Q_i) = \frac{|C(Q_i) \cap C^*|}{|C^*|} \quad \in [0, 1].
\end{equation}

Alternatively, for finer granularity, we compute {Jaccard similarity} over {normalized question skeletons} (replacing named entities with \texttt{[ENT]}, numbers with \texttt{[NUM]}, etc.) to reward structural diversity beyond category labels. \( R_{\text{Structure}} \) is assigned the {highest weight} (\( w_2 \gg w_1, w_3 \)) to prioritize comprehensive evidence gathering.

\vspace{2mm}
\noindent
\textbf{Accuracy Reward (\( R_{\text{Accuracy}}(E_i) \))}.  
This reward measures factual relevance of the retrieved evidence \( E_i = \text{Retrieve}(Q_i, G) \) to the claim \( c \). We use a binary signal from an LLM judge (e.g., Qwen2.5-72B-Instruct) or a fine-tuned factuality classifier:  

- Input: Concatenated evidence $( E_i )$ and claim $( c)$.

- Output: Label \texttt{SUPPORTS}, \texttt{REFUTES}, or \texttt{NEI}. 
The reward is:  
\begin{equation}
    R_{\text{Accuracy}}(E_i) = 
\begin{cases} 
1 & \text{if information is correct,} \\
0 & \text{if hallucinated}.
\end{cases}
\end{equation}
 
By combining these three rewards, GRPO creates a hierarchical optimization landscape: the model is guided primarily by structure to explore the claim comprehensively, constrained by format for reliability, and refined by accuracy for verdict quality. This yields question sets that are well-formed, semantically rich, and evidentially potent, critical for robust misinformation detection.

\section{Experiment and Results}

\subsection{Experimental Setup}

\subsubsection{Datasets}

\label{sec:dataset}

To evaluate our framework, we utilize three benchmark datasets for fake news detection, each with distinct characteristics to assess evidence retrieval, verdict accuracy, and explanation quality:

\begin{itemize}
    \item \textbf{PolitiFact} \citep{shu2020fakenewsnet}: Contains political claims from U.S. media, annotated with veracity labels (e.g., True, False, Pants on Fire) and detailed justifications, ideal for testing verdict accuracy and explanation coherence.
    \item \textbf{LIAR} \citep{wang2017liar}: Comprises 12,8K short political statements from diverse sources, labeled with fine-grained veracity categories (e.g., True, Mostly True, False), suitable for evaluating generalization across varied claims.
    \item \textbf{FEVER} \citep{thorne2018fever}: Includes Wikipedia-derived claims, annotated as Supported, Refuted, or Not Enough Information, with linked evidence, enabling robust assessment of evidence retrieval and multi-hop reasoning.
\end{itemize}

These datasets collectively challenge our framework’s ability to handle diverse claim types, structured evidence, and contextual nuances in misinformation detection, as summarized in Table \ref{tab:dataset-counts}.
\begin{table}[t]
    \centering
    \caption{Counts of real/fake news across datasets.}
    \label{tab:dataset-counts}
    \begin{tabular}{l S S S}
        \toprule
        & \multicolumn{1}{c}{LIAR} & \multicolumn{1}{c}{{FEVER}} & \multicolumn{1}{c}{POLITIFACT} \\
        \midrule
        \#Real News & 9252 & 3333 & 399 \\
        \#Fake News & 3555 & 3333 & 345 \\
        \#Total     & 12807 & 6666 & 744 \\
        \bottomrule
    \end{tabular}
\end{table}

\subsubsection{Baseline}

To benchmark our \theName\  framework, we compare it against five methods for fake news detection, each highlighting different retrieval and reasoning capabilities:

\begin{itemize}
    \item \textbf{Vanilla LLM} \citep{ji2023survey}: Uses a pretrained large language model for direct claim classification without external knowledge, prone to hallucinations in verification tasks.
    \item \textbf{Naive RAG} \citep{lewis2020retrieval}: Performs one-step retrieval from an unstructured corpus to augment prompts, often retrieving irrelevant or incomplete evidence due to lack of query refinement.
    \item \textbf{LightRAG} \citep{guo2024lightrag}: Employs graph-enhanced indexing for faster, contextual retrieval, reducing computational overhead while maintaining accuracy in knowledge-intensive queries.
    \item \textbf{ReAct} \citep{yao2023react}: Interleaves reasoning and acting via chain-of-thought and tool calls (e.g., search APIs), enabling dynamic evidence gathering for multi-hop fact-checking.
    \item \textbf{HippoRAG2} \citep{jimenez2024hipporag}: Builds hierarchical knowledge graphs with personalized PageRank for incremental, context-aware retrieval, excelling in integrating diverse evidence.
\end{itemize}

\subsubsection{Implementation Details}

Our \theName\  framework utilizes GPT-4 for constructing the Knowledge Graph from verified datasets, leveraging its advanced language understanding for entity and relationship extraction \citep{achiam2023gpt}. For the question generation and verdict prediction, we benchmark three scales of Qwen2.5 (1.5B, 3B, and 7B parameters), evaluating their performance across diverse claim complexities \citep{team2024qwen2}. The retrieval module employs the bge-large-en-v1.5 model for embedding questions and KG elements, ensuring robust similarity-based evidence extraction \citep{chen2024bge}. 

GRPO Training Configuration: We implement GRPO optimization using a group size of \(N = 8\) question sets per claim for reliable advantage estimation. The reward function weights are set as \(w_1 = 0.15\), \(w_2 = 0.60\), and \(w_3 = 0.25\), where \(w_2 \gg w_1, w_3\) to prioritize structural diversity as described in Section 4.6. We use a clipping parameter \(\epsilon = 0.20\) and KL penalty coefficient \(\beta = 0.04\) to ensure stable policy updates. Training is performed on four NVIDIA A100 GPUs (80GB) with a learning rate of \(5 \times 10^{-6}\) and global batch size of 32 claims. For Qwen2.5-7B, GRPO fine-tuning converges in approximately 10.5 hours, while the 1.5B and 3B variants require 3.0 and 5.5 hours respectively. Complete hyperparameters and KG statistics are detailed in Table \ref{tab:implementation_details}.

\begin{table}[htbp]
\centering
\scriptsize
\caption{DeLIVeR Training Hyperparameters and Knowledge Graph Statistics}
\label{tab:implementation_details}
\begin{tabular}{lcc}
\toprule
\textbf{Parameter} & \textbf{Setting/Value} & \textbf{Notes} \\
\midrule
\multicolumn{3}{c}{\textbf{GRPO Hyperparameters}} \\
\midrule
Group size ($N$) & 8 & Question sets sampled per claim \\
Format reward weight ($w_1$) & 0.15 & Binary format validation \\
Structure reward weight ($w_2$) & 0.60 & Semantic diversity priority ($w_2 \gg w_1, w_3$) \\
Accuracy reward weight ($w_3$) & 0.25 & Evidence relevance to verdict \\
Clipping parameter ($\epsilon$) & 0.20 & PPO-style policy update stability \\
KL penalty coefficient ($\beta$) & 0.04 & Reference model regularization \\
Learning rate & $5 \times 10^{-6}$ & AdamW optimizer \\
Warmup steps & 100 & Linear learning rate warmup \\
Global batch size & 32 & Claims per gradient update \\
Max training epochs & 3 & Early stopping on validation F1 \\
Temperature (sampling) & 0.7 & Diversity in group generation \\
\midrule
\multicolumn{3}{c}{\textbf{Computational Requirements}} \\
\midrule
Hardware & 4 $\times$ NVIDIA A100 (80GB) & Distributed training setup \\
Training time (Qwen2.5-1.5B) & $\sim$3.0 hours & Average across datasets \\
Training time (Qwen2.5-3B) & $\sim$5.5 hours & Average across datasets \\
Training time (Qwen2.5-7B) & $\sim$10.5 hours & Average across datasets \\
KG construction time & $\sim$6.0 hours & GPT-4 OpenIE extraction \\
Peak GPU memory (7B model) & $\sim$65 GB & Including gradients and optimizer \\
\midrule
\multicolumn{3}{c}{\textbf{Knowledge Graph Statistics}} \\
\midrule
LIAR: Entities / Triples & 42,150 / 128,400 & Political claims and entities \\
FEVER: Entities / Triples & 186,300 / 524,800 & Wikipedia-derived evidence \\
PolitiFact: Entities / Triples & 8,920 / 26,150 & Fact-checking corpus \\
Avg. entity connectivity & 3.2 & Triples per entity (post-filtering) \\
Relation types (total) & 47 & Unique predicate categories \\
Singleton filter threshold & $\geq 2$ connections & Minimum connectivity requirement \\
Temporal edge coverage & 58.7\% & Edges with timestamp metadata \\
\bottomrule
\end{tabular}
\end{table}

\subsection{ Main Results}

Table \ref{tab:main_result} presents a comprehensive evaluation of our \theName\  framework against five baselines include Vanilla LLM, Naive RAG, LightRAG, ReAct, and HippoRAG2 across the LIAR, FEVER, and PolitiFact datasets, using three scales of the Qwen2.5 model (1.5B, 3B, and 7B parameters). Our framework consistently outperforms all baselines across recall, precision, accuracy, and F1-score, achieving peak F1-scores of 83.73 (LIAR), 84.57 (FEVER), and 79.70 (PolitiFact) with Qwen2.5-7B, compared to HippoRAG2’s best F1-scores of 72.93, 75.16, and 69.97, respectively. This represents an average F1 improvement of 10–15\% over HippoRAG2, the closest competitor, highlighting the efficacy of our knowledge graph (KG) integration and Group Relative Policy Optimization (GRPO)-driven question set generation. The performance gap widens with larger model scales, with Qwen2.5-7B yielding the highest scores due to its enhanced reasoning capacity, which better leverages the structured evidence retrieved from the KG. FEVER consistently produces the highest scores across all methods, likely due to its structured Wikipedia-based evidence annotations, which align well with our framework’s multi-hop reasoning capabilities. In contrast, PolitiFact’s smaller dataset size (744 samples) and nuanced veracity labels (e.g., Pants on Fire) pose greater challenges, yet our framework still achieves robust performance (F1 of 79.70 with Qwen2.5-7B), demonstrating its ability to handle complex, real-world misinformation scenarios.

\begin{table*}[t!]
\centering
\scriptsize
\caption{Performance Metrics on MultiReQA Datasets}
\label{tab:main_result}
\resizebox{\textwidth}{!}{
\renewcommand{\arraystretch}{1.05}
\begin{tabular}{l|cccc|cccc|cccc}
\toprule
\multirow{4}{*}{\textbf{Methods}}

& \multicolumn{4}{c|}{\textbf{LIAR}} & \multicolumn{4}{c|}{\textbf{FEVER}} 
& \multicolumn{4}{c}{\textbf{POLITIFACT}} \\
\cmidrule(lr){2-5} \cmidrule(lr){6-9} \cmidrule(lr){10-13}

& \multicolumn{12}{c}{\rule{0pt}{1.2em}\textbf{\textit{Qwen2.5-1.5B-Instruct}}\rule[-1.0em]{0pt}{1.2em}} \\[2pt]
& Recall & Prec & Acc & F1 & Recall & Prec & Acc & F1 & Recall & Prec & Acc & F1 \\
\midrule
Vanilla LLM & 36.51 & 35.37 & 35.92 & 35.93 & 37.64 & 36.22 & 36.59 & 36.89 & 35.48 & 34.91 & 34.78 & 35.19 \\
Naive RAG & 52.47 & 49.36 & 50.41 & 50.87 & 54.12 & 51.59 & 52.31 & 52.83 & 50.94 & 48.23 & 49.32 & 49.63 \\
LightRAG & 58.64 & 55.22 & 56.15 & 56.89 & 60.53 & 57.18 & 58.02 & 58.83 & 57.25 & 54.37 & 55.18 & 55.74 \\
ReAct & 60.18 & 57.34 & 58.10 & 58.72 & 62.81 & 59.47 & 60.32 & 61.11 & 61.29 & 58.42 & 59.30 & 59.82 \\
HippoRAG2 & 63.72 & 60.35 & 61.44 & 61.96 & 65.33 & 62.24 & 63.15 & 63.77 & 63.61 & 60.23 & 61.04 & 61.86 \\
\rowcolor{blue!20} \textit{\theName\ (ours)} & \textbf{72.48} & \textbf{69.53} & \textbf{70.31} & \textbf{70.97} & \textbf{74.69} & \textbf{71.42} & \textbf{72.24} & \textbf{72.99} & \textbf{70.18} & \textbf{67.23} & \textbf{68.11} & \textbf{68.68} \\
\midrule

& \multicolumn{12}{c}{\rule{0pt}{1.2em}\textbf{\textit{Qwen2.5-3B-Instruct}}\rule[-1.0em]{0pt}{1.2em}} \\[2pt]
\midrule
Vanilla LLM & 38.17 & 37.02 & 37.56 & 37.58 & 39.45 & 38.13 & 38.42 & 38.79 & 37.68 & 36.59 & 36.74 & 37.13 \\
Naive RAG & 56.41 & 53.72 & 54.38 & 55.03 & 58.29 & 55.36 & 56.02 & 56.80 & 55.72 & 52.64 & 53.32 & 53.98 \\
LightRAG & 62.68 & 59.39 & 60.15 & 60.97 & 64.43 & 61.08 & 61.93 & 62.77 & 61.89 & 58.66 & 59.48 & 60.24 \\
ReAct & 65.33 & 62.47 & 63.18 & 63.86 & 67.28 & 64.10 & 65.09 & 65.63 & 64.47 & 61.52 & 62.34 & 62.88 \\
HippoRAG2 & 69.28 & 66.14 & 67.09 & 67.69 & 71.54 & 68.22 & 69.13 & 69.85 & 67.88 & 64.61 & 65.54 & 66.28 \\
\rowcolor{blue!20} \textit{\theName\ (ours)} & \textbf{79.37} & \textbf{76.44} & \textbf{77.32} & \textbf{77.88} & \textbf{81.25} & \textbf{78.16} & \textbf{79.03} & \textbf{79.67} & \textbf{75.73} & \textbf{72.66} & \textbf{73.49} & \textbf{73.98} \\
\midrule

& \multicolumn{12}{c}{\rule{0pt}{1.2em}\textbf{\textit{Qwen2.5-7B-Instruct}}\rule[-1.0em]{0pt}{1.2em}} \\[2pt]
\midrule
Vanilla LLM & 39.23 & 38.15 & 38.42 & 38.69 & 40.51 & 39.29 & 39.48 & 39.90 & 38.74 & 37.59 & 37.74 & 38.16 \\
Naive RAG & 60.44 & 57.32 & 58.21 & 58.85 & 62.73 & 59.48 & 60.26 & 60.98 & 60.92 & 57.18 & 58.09 & 58.61 \\
LightRAG & 66.38 & 63.20 & 64.12 & 64.76 & 68.44 & 65.19 & 66.07 & 66.78 & 64.69 & 61.58 & 62.43 & 63.09 \\
ReAct & 70.52 & 67.33 & 68.28 & 68.87 & 72.86 & 69.61 & 70.43 & 71.18 & 69.27 & 66.13 & 67.04 & 67.64 \\
HippoRAG2 & 74.63 & 71.35 & 72.29 & 72.93 & 76.85 & 73.52 & 74.41 & 75.16 & 71.66 & 68.39 & 69.33 & 69.97 \\
\rowcolor{blue!20} \textit{\theName\ (ours)} & \textbf{85.32} & \textbf{82.21} & \textbf{83.09} & \textbf{83.73} & \textbf{86.14} & \textbf{83.11} & \textbf{84.00} & \textbf{84.57} & \textbf{81.36} & \textbf{78.23} & \textbf{79.18} & \textbf{79.70} \\
\bottomrule
\end{tabular}
}
\end{table*}

\subsection{Ablation Study}
\subsubsection{Impact of question set size: }

\begin{figure}[htbp]
\centering
\includegraphics[width=0.5\linewidth]{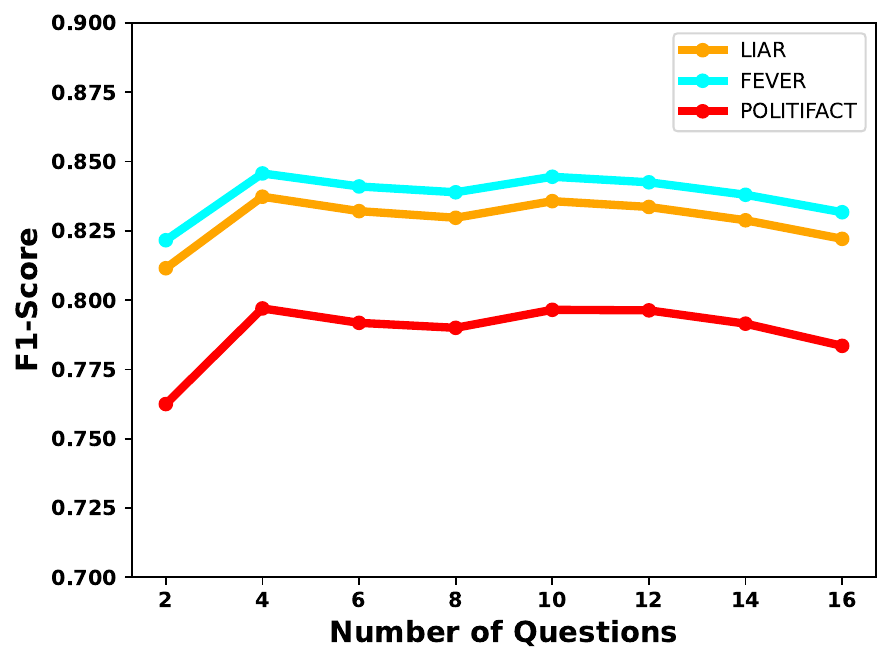}

\caption{The impact of the number of generated questions}
\label{fig:Number_of_Questions}
\end{figure}

We ablate question set size using Qwen2.5-7B-Instruct on LIAR, FEVER, and PolitiFact (Figure \ref{fig:Number_of_Questions}). Peak F1-scores are achieved with four questions: 0.8373 (LIAR), 0.8457 (FEVER), and 0.7970 (PolitiFact), confirming the optimal balance between evidence coverage and relevance. Performance declines with eight questions (0.8297, 0.8389, 0.7900) due to redundancy and with sixteen questions (0.8221, 0.8317, 0.7835) due to retrieval noise. Only two questions yield the lowest scores (0.8115, 0.8216, 0.7625) from insufficient evidence diversity. FEVER benefits most from structured evidence, while PolitiFact remains the hardest due to nuanced labels and limited data. These results validate GRPO’s effectiveness in producing concise, high-quality question sets and underscore the importance of tuning set size for robust retrieval and verdict accuracy in misinformation detection.

\subsubsection{Qualitative Evaluation}
\begin{figure}[htbp]
\centering
\includegraphics[width=0.8\linewidth]{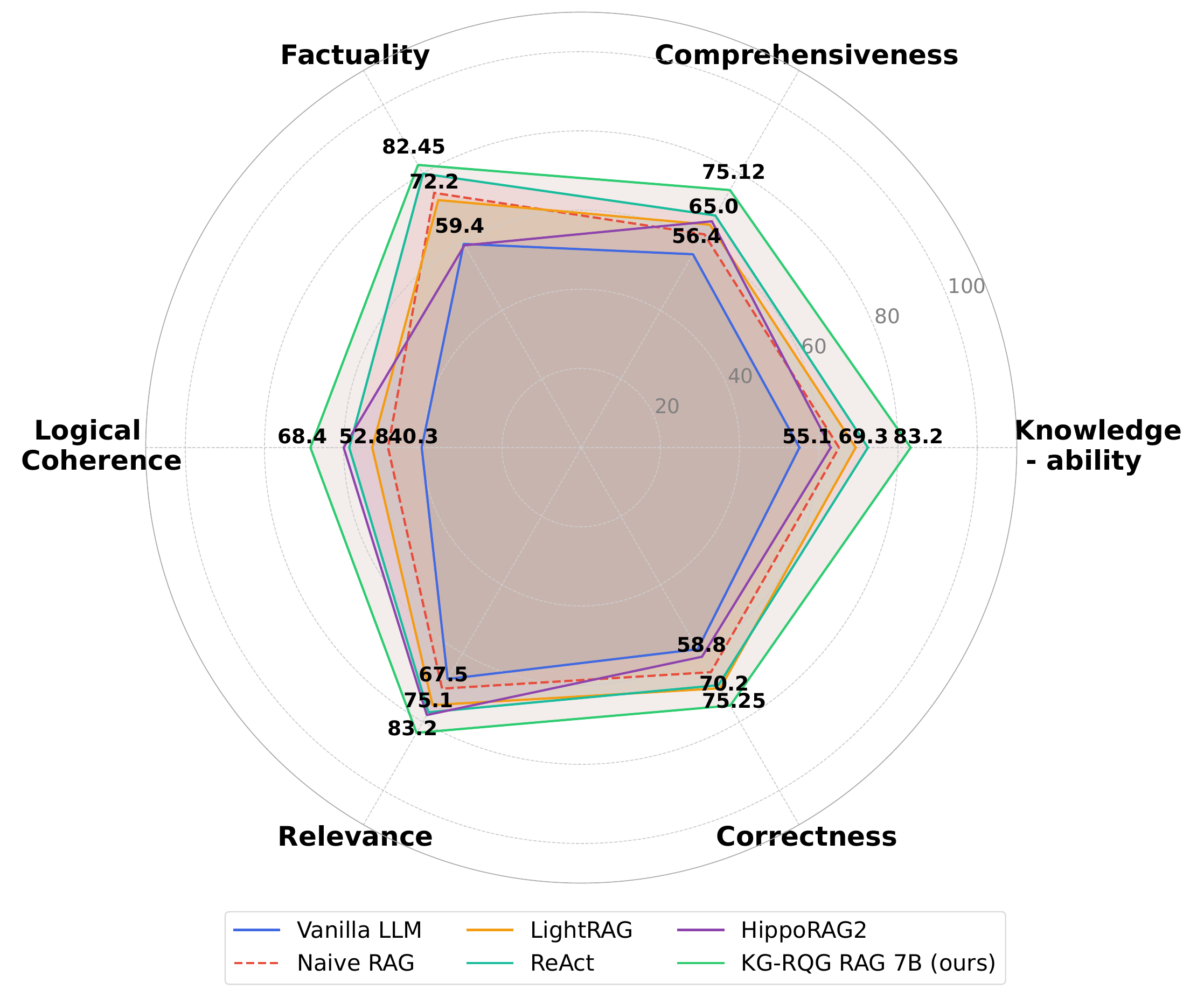}

\caption{Qualitative performance of our \theName\ framework in six categories.}
\label{fig:qualitative}
\end{figure}

Figure \ref{fig:qualitative} illustrates the qualitative performance of our \theName\ framework (Qwen2.5-7B-Instruct) against Vanilla LLM, Naive RAG, LightRAG, ReAct, and HippoRAG2 across six categories for misinformation detection: Knowledge-ability, Comprehensiveness, Factuality, Logical Coherence, Relevance, and Correctness. Evaluated on LIAR, FEVER, and PolitiFact, our framework achieves top scores: 83.2 (Knowledge-ability, Relevance), 75.12 (Comprehensiveness), 82.45 (Factuality), 68.4 (Logical Coherence), and 75.25 (Correctness), outperforming ReAct’s 72.4, 67.7, 79.9, 58.6, 77.2, and 69.3, respectively. KG-driven retrieval and GRPO-optimized questions enhance Factuality and Relevance, while Vanilla LLM struggles (e.g., 40.3 in Logical Coherence) due to limited knowledge. Naive RAG (65.1 in Knowledge-ability), LightRAG (69.3), and HippoRAG2 (59.0 in Factuality) lag behind. These results validate our framework’s robust, coherent verdicts for misinformation detection.

\begin{table}[h]
\centering
\footnotesize
\label{tab:error_types}
\begin{tabular}{l c}
\hline
\textbf{Error Types} & \textbf{\theName} \\
\hline
Irrelevant Questions     & 16\% \\
Insufficient Coverage    & 34\% \\
Redundant Questions      & 14\% \\
Document Mismatch        & 48\% \\
\hline
\end{tabular}
\vspace{5mm}
\caption{Distribution of errors based on 200 examples from POLITIFACT, where \theName\ gives incorrect verification results.}
\end{table}

\subsubsection{Error analysis of the retrieval process: }

We conduct error analysis on 200 PolitiFact failure cases where \theName\ predicts incorrect verdicts. Manual annotation reveals four error types (Table \ref{tab:error_types}): Document Mismatch (48\%) dominates, indicating poor alignment between questions and retrieved evidence, especially for nuanced or temporally sensitive claims. Insufficient Coverage (34\%) ranks second, showing that critical aspects (e.g., motive, source credibility) are sometimes missed, particularly in “Mostly False”/“Half-True” claims. Encouragingly, Irrelevant Questions (16\%) and Redundant Questions (14\%) together account for only 30\%, confirming that GRPO effectively generates focused, non-repetitive questions. This supports our ablation finding that four questions achieve optimal balance. These results highlight the need to (1) improve question-evidence alignment via retriever fine-tuning and (2) strengthen GRPO rewards to penalize Document Mismatch more heavily. Addressing these will significantly enhance retrieval-augmented fact-checking robustness.

\section{Conclusion}

We presented a fact-verification architecture that integrates knowledge graph grounding and targeted, set-based question planning on top of a frozen verifier trained through GRPO to monotonically improve accuracy, interpretability, and robustness over LLM-only and recent RAG baselines. By decomposing claims into focused questions, retrieving precise multi-hop evidence, and rationalizing a True/False/NEI decision with user-aligned feedback, our approach produces fully auditable reasoning traces suitable for high-stakes applications such as newsrooms, content moderation teams, election monitoring, public health communication, and enterprise compliance workflows.
Unlike black-box LLM judgments or noisy single-shot RAG retrievals, our method exposes every inference step from generated questions to retrieved KG paths to final verdict—enabling human auditors to verify or contest any component. This transparency is critical when decisions influence policy, public perception, or legal outcomes. We show that the combination of structured retrieval and reinforced question generation narrows the gap between raw model power and real-world trust requirements, achieving state-of-the-art verdict accuracy while delivering human-readable explanations at scale. Ultimately, this work advances toward deployable, accountable AI systems capable of combating misinformation with both rigor and responsibility.

\section{Acknowledgment}

This work was supported by NSF - USA CNS-2219614.

\end{document}